\documentclass[fleqn,10pt]{wlscirep}
\usepackage[utf8]{inputenc}
\usepackage[T1]{fontenc}
\usepackage{lineno}
\usepackage{colortbl}
\usepackage{hyperref}
\usepackage{ifthen}

\newboolean{revised}
\setboolean{revised}{false}

\ifthenelse{\boolean{revised}}
{

\usepackage{xcolor}
\usepackage[normalem]{ulem}
\setlength{\marginparwidth}{\dimexpr\marginparwidth+0.1cm\relax}
\setlength{\marginparsep}{\dimexpr\marginparsep+0.05cm\relax}
\usepackage{marginnote}
\usepackage{xspace}

\newcommand{\revref}[2]{%
\marginnote{$\scriptstyle {R_{#1}C_{#2}} $}[-0.36cm]
}

\DeclareRobustCommand{\robustrevref}[2]{%
\revref{#1}{#2}
}

\newcommand{\revmod}[1]{%
{\color{blue}#1}\xspace
}

\newcommand{\revnew}[1]{%
{\color{orange}#1}\xspace
}

\newcommand{\revdel}[1]{%
{\color{darkgray}\sout{#1}}\xspace
}

}{
    \usepackage{xspace}
    \newcommand{\marginnote}[1]{\ignorespaces}
    \newcommand{\revref}[2]{\ignorespaces}
    \newcommand{\revmod}[1]{#1\xspace}
    \newcommand{\revnew}[1]{#1\xspace}
    \newcommand{\revdel}[1]{\ignorespaces}
    \newcommand{\robustrevref}[2]{\ignorespaces}
}

\newcommand{\figref}[1]{\figurename~\ref{#1}}
\newcommand{\tabref}[1]{\tablename~\ref{#1}}

\newcommand{\greencheck}{{\color{green}\checkmark}}

\title{CholecInstanceSeg: A Tool Instance Segmentation Dataset for Laparoscopic Surgery}

\author[1]{Oluwatosin Alabi}
\author[2]{Ko Ko Zayar Toe}
\author[4]{Zijian Zhou}
\author[1]{Charlie Budd}
\author[1]{Nicholas Raison}
\author[3,*]{Miaojing Shi}
\author[1]{Tom Vercauteren}
\affil[1]{Kings College London,  Surgical \& Interventional Engineering, London, SE1 7EU
, United Kingdom}
\affil[2]{Kings College Hospital Denmark Hill, department, London, SE5 9RS, United Kingdom}
\affil[3]{Tongji University, College of Electronic and Information Engineering, Shanghai, 200092, China}
\affil[4]{Department of Informatics, King’s College London}

\affil[*]{Corresponding author}

\begin{abstract}
In laparoscopic and robotic surgery, precise tool instance segmentation is an essential technology for advanced
computer-assisted interventions.
Although publicly available procedures of routine surgeries exist, they often lack comprehensive annotations for tool instance segmentation.
Additionally, the majority of standard datasets for tool segmentation are derived from porcine(pig) surgeries.
To address this gap, we introduce CholecInstanceSeg, the largest open-access tool instance segmentation dataset to date. 
Derived from the existing CholecT50 and Cholec80 datasets, CholecInstanceSeg provides novel annotations for laparoscopic cholecystectomy procedures in patients.
Our dataset comprises 41.9k annotated frames extracted from 85 clinical procedures and 64.4k tool instances, each labelled with semantic masks and instance IDs.
To ensure the reliability of our annotations, we perform extensive quality control, conduct label agreement statistics, and benchmark the segmentation results with various instance segmentation baselines. 
CholecInstanceSeg aims to advance the field by offering a comprehensive and high-quality open-access dataset for the development and evaluation of tool instance segmentation algorithms. 
\revdel{CholecInstanceSeg is publicly available \href{https://www.synapse.org/Synapse:syn60239970/wiki/628710}{here}}
\end{abstract}
\begin{document}

\flushbottom
\maketitle

\thispagestyle{empty}


\section*{Background \& Summary}
Minimally invasive surgeries (MIS) have gained popularity due to their benefits over open surgeries, such as reduced blood loss, less pain, faster recovery times, and fewer post-surgical complications~\cite{fuchs2002minimally}. However, MIS relies on an endoscope or laparoscope for indirect vision, which provides a limited field of view, making it challenging for surgeons to accurately interpret complex surgical scenes ~\cite{real_time_seg_adverserial}. To overcome this, computer-aided intervention techniques that enhance the information obtained through minimally invasive cameras have been proposed~\cite{vecauteren_cai_4_cai,cv_in_surgery}.
At the forefront of this evolution lies the challenge of achieving precise instance segmentation of surgical tools in a minimally invasive surgical scene.
This task is essential for developing assistive surgical technology and autonomous or semi-autonomous surgical systems, which require detailed knowledge of the identification and localization of various tools.

Current publicly available datasets for surgical tool segmentation have several notable limitations.
Many focus on porcine or ex-vivo surgeries rather than routine human procedures, limiting their applicability to day-to-day clinical settings~\cite{endovis2015,endovis2017,endovis2018}.
Others are multitask datasets with limited tool segmentation data~\cite{autolaparo}, or they provide only binary instance segmentation for a single tool class~\cite{robustmis2019}. Additionally, available surgical tool segmentation datasets lack instance-specific information ~\cite{endovis2015,endovis2017,endovis2018,autolaparo,psychogyios2023sar,cholecseg8k}, and most offer a relatively small number of annotated segmentation masks.

\revdel{To overcome these challenges, we introduce the CholecInstanceSeg open-access dataset, in which we meticulously annotate 41,933 frames extracted from 85 laparoscopic cholecystectomy procedures from the existing Cholec80 \mbox{~\cite{endonet}}
 and CholecT50 \mbox{~\cite{NWOYE2022102433_rendevous}} datasets.}

To overcome these challenges, we introduce the CholecInstanceSeg open-access dataset, in which we meticulously annotate 41,933 frames from the existing Cholec80~\cite{endonet} and CholecT50~\cite{NWOYE2022102433_rendevous} datasets \robustrevref{1}{2}\revnew{with instance segmentation labels and class labels for 7 instrument classes: Grasper, Bipolar, Hook, Clipper, Scissors, Irrigator, and Snare. 
CholecInstanceSeg contains frames from 85 videos of Laparoscopic cholecystectomy - a routine surgical procedure for gallbladder removal, commonly performed to treat gallstones or cholecystitis with over 300,000 surgeries performed annually in the United States \cite{hassler2023laparoscopic}}.
To the best of our knowledge,
CholecInstanceSeg represents the most extensive surgical tool instance segmentation dataset.
\tabref{tab:compare_datasets} compares popular datasets used in tool segmentation research and CholecInstanceSeg. We acknowledge a concurrent effort, \href{https://phakir.re-mic.de/data/}{the PhaKIR dataset}, which is being developed for the upcoming MICCAI2024 PhaKIR-challenge. The PhaKIR dataset is expected to include 15 videos and approximately 30,000 instance annotations. However, as this dataset is only partially released and not yet available for research, we have not included it in our comparisons.

\begin{table}[ht]
  \centering
  \begin{tabular}{|l|l|l|l|l|l|l|} 
    \hline
     \textbf{Dataset} & \textbf{Patient} & \textbf{Environment} & \textbf{Procedures} & \textbf{Annotations} &  \textbf{Semantic} & \textbf{Instance} \\
    \hline
    Endovis2015 \cite{endovis2015}  & Human & Ex-vivo and In-vivo & 8 & 10k &  3 classes  & No \\
    \hline
    Endovis2017 \cite{endovis2017} & Porcine & In-vivo  & 10 & 2.4k  & 10 classes  & No  \\
    \hline
    Endovis2018 \cite{endovis2018} & Porcine & In-vivo & 16 & 2.4k  &  10 classes & No  \\
    \hline
    CholecSeg8k \cite{cholecseg8k} & Human & In-vivo & 17 &  8k   & 12 classes&  No \\
    \hline
    AutoLaparo (Multitask) \cite{autolaparo} & Human & In-vivo & 21 & 1.8k  & 7 classes  & No   \\
    \hline
    ROBUSTMIS2019 \cite{robustmis2019} & Human & In-vivo & 30 &  10k   & 2 classes & Yes\greencheck \\
    \hline
    SAR-RARP50 (Multitask) \cite{psychogyios2023sar} & Human & In-vivo & 50  & 10k  & 10 classes  & No \\ 
    \hline
    CholecInstanceSeg (Ours) & Human & In-vivo & 85 & 41.9k & 8 classes & Yes\greencheck   \\
    \hline
  \end{tabular}
  \caption{Comparison of popular datasets used in tool segmentation research, indicating the type of surgeries, the number of surgical procedures, the number of frames annotated, the semantic classes annotated, and whether each dataset contains instance information.}
  \label{tab:compare_datasets}
\end{table}

Furthermore, we provide technical validations of our dataset, demonstrating our comprehensive quality control procedures and showcasing CholecInstanceSeg's capability for training instance segmentation models.

CholecInstanceSeg is ideally suited for training models on instance and semantic segmentation for laparoscopic surgeries. Additionally, it can be effectively integrated with other datasets that utilize surgical procedure videos from Cholec80 and CholecT50. These datasets, derived from Cholec80 and CholecT50 share frames with CholecInstanceSeg but are annotated for different tasks, such as phase recognition (Cholec80 \cite{endonet}) instrument action and target detection (CholecT50 \cite{NWOYE2022102433_rendevous}), and Critical View of Safety (CVS) assessment (Cholec80-CVS \cite{rios2023cholec80}).

\section*{Methods}
This section outlines the methodology for creating the CholecInstanceSeg dataset, divided into four main topics: data sources, labelling protocol, annotation tool and techniques, and the annotation workflow and procedures.

First, we describe the data sources and how the frames were extracted. Next, we outline the principles guiding our instance segmentation dataset, emphasizing the labelling protocol. We then discuss the tools and techniques used to overcome annotation challenges. Finally, we provide a detailed account of the annotation process, including information about the annotators and the time spent on creating CholecInstanceSeg.

\subsection*{Data Sources}
\begin{figure}[ht]
\centering
\includegraphics[width=\linewidth]{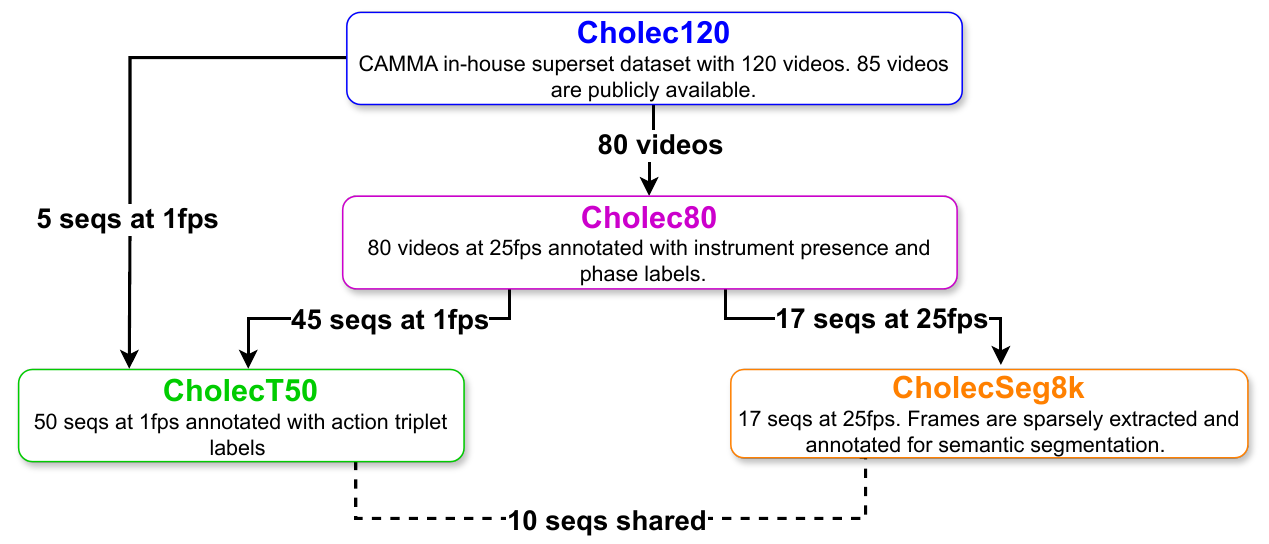}
\caption{CholecInstanceSeg contains frames extracted from CholecT50, CholecSeg8k, and Cholec80. These three datasets are sub-datasets of the CAMMA in-house dataset called Cholec120. There are also some shared image sequences (referred to as \emph{seqs} in the figure for brevity) between CholecT50 and CholecSeg8k.}
\label{fig:cholec_dataset_relationships}
\end{figure}

The CholecInstanceSeg dataset is curated using frames extracted from three existing laparoscopic cholecystectomy datasets: Cholec80 \cite{endonet}, CholecT50 \cite{NWOYE2022102433_rendevous}, and CholecSeg8k \cite{cholecseg8k}, containing a total of 85 image sequences. 
These datasets are publicly available subsets of the in-house Cholec120 dataset \cite{twinanda2018rsdnet} from the CAMMA research group \robustrevref{1}{7}\revnew{which was recorded at the University Hospital of Strasbourg \cite{endonet,NWOYE2022102433_rendevous}} 
. The interrelation between these datasets is illustrated in  \figref{fig:cholec_dataset_relationships}. 

When discussing the source datasets, we distinguish between videos and image sequences. Videos are multimedia compressed files provided in formats such as .mp4, while image sequences refer to extracted sequential frames from videos. 

\robustrevref{1}{5}\revnew{The frames from Cholec80 and CholecT50 used in CholecInstanceSeg are primarily at a resolution of 854x480 pixels, consistent with the original Cholec80 and CholecT50 datasets. However, three sequences from CholecT50 are provided at a higher resolution of 1920x1080 pixels. CholecT50 contains pre-extracted image sequences and we used these as is, while frames from Cholec80 were extracted at 1 fps using the FFMPEG library.}

CholecInstanceSeg includes frames from publicly available datasets that already have the necessary consents and ethical considerations. Additionally, the licenses for the source datasets permit the enhancement of the data with further annotations (\href{https://github.com/CAMMA-public/TF-Cholec80}{Cholec80 license}, \href{https://github.com/CAMMA-public/cholect50}{CholecT50 license}, \href{https://www.kaggle.com/datasets/newslab/cholecseg8k}{CholecSeg8k license}).

\subsubsection*{CholecSeg8k}
CholecSeg8k comprises 8,080 laparoscopic cholecystectomy image frames extracted from 17 surgical procedure video clips within the Cholec80 dataset. This dataset provides 80 consecutive images sparsely annotated for semantic segmentation (but not instance segmentation).
To enhance CholecSeg8k, instance IDs are required for the existing semantic segmentation labels. As CholecSeg8k already contains semantic segmentation labels, it serves as a good starting point for further annotation. However, it is important to note that CholecSeg8k includes only two tool classes: Grasper and Hook, which is fewer than the number of tool classes  in the parent datasets.

\subsubsection*{CholecT50}
CholecT50 is a publicly available dataset comprising 50 real-world videos of laparoscopic cholecystectomy surgeries. This dataset integrates 45 videos from the Cholec80 dataset along with 5 additional videos from the in-house Cholec120 dataset. 
The image sequences in CholecT50 are extracted at a rate of 1 frame per second (fps). Each frame in CholecT50 is annotated with fine-grained tool-tissue interaction labels, specifying the \texttt{<instrument, action, target>}. Although CholecSeg8k and CholecT50 share 10 image sequences, we confirmed that the video extraction methods used for these datasets are different. We checked the similarity between extracted frames in the shared sequences from both datasets and found no exact matches.

The CholecT50 dataset consists of 100.9k frames distributed across 50 image sequences. Given the high cost of annotating every single frame, we adopted a strategy to balance thoroughness and feasibility. Consequently, we divided CholecT50 into two partitions:
\begin{enumerate}
    \item CholecT50-full: This partition was chosen to feature comprehensive annotations, including instance IDs and segmentation masks for every frame in the image sequence. This thorough annotation provides detailed data for in-depth analysis and model training, allowing for the exploration of ideas like video segmentation.
    \item CholecT50-sparse: In this partition, annotations are provided sparsely in time, yet they cover the entire sequence. This approach reduces annotation costs while still offering valuable insights into the location and identification of tools for each sequence. We sample one in every 30 frames ($\frac{1}{30}$ fps) and add more frames if the resulting sample contains fewer than 50 frames per sequence.
\end{enumerate}

\subsubsection*{Cholec80}
Cholec80 comprises 80 videos of cholecystectomy surgeries. It is labelled with phase information at 25 fps and tool presence annotations at 1 fps. The tool annotations include instruments where at least half of the tool-tip is visible. There are six tools available in the provided tool presence annotations.

Given the overlap between Cholec80, CholecSeg8k, and CholecT50, we focus on annotating the 28 videos in Cholec80 that are not present in CholecSeg8k and CholecT50. We extract frames from these 28 videos at a rate of $\frac{1}{30}$ fps and provide annotations for these frames. 
To maintain consistency with the extraction protocol of CholecT50 and CholecSeg8k, we blacked out frames that showed the exterior of the body via the laparoscope. Additionally, if the resulting sample contained fewer than 50 frames per sequence, more frames were included to ensure adequate coverage.

\subsubsection*{Summary of CholecInstanceSeg Partitions}
\begin{figure}[ht]
  \centering
  \includegraphics[width=\linewidth]{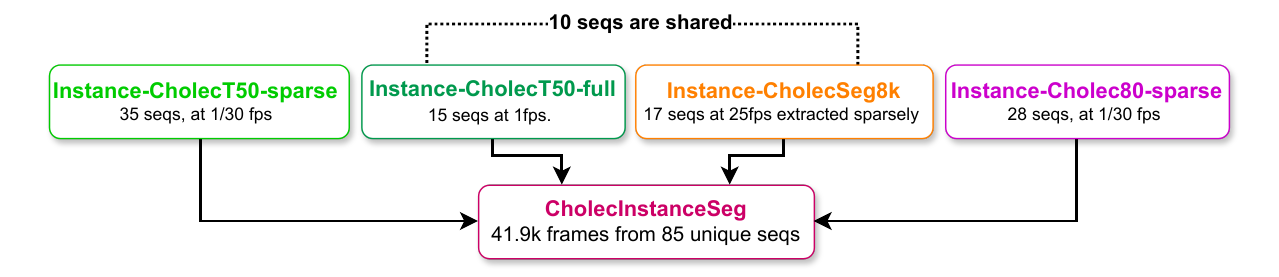}
   \caption{CholecInstanceSeg is composed of frames from four dataset partitions: Instance-CholecT50-sparse, Instance-CholecT50-full, Instance-CholecSeg8k, and Instance-Cholec80-sparse. The diagram illustrates the number of image sequences (denoted \emph{seqs} in the figure for brevity) and extraction rates for each partition. Note that 10 sequences are shared between Instance-CholecT50-full and Instance-CholecSeg8k.}
   \label{fig:our_dataset} 
\end{figure}

CholecInstanceSeg, which contains 41.9k frames from 85 unique image sequences, can be partitioned into four distinct sections based on the data source: Instance-CholecSeg8k, Instance-CholecT50-full, Instance-CholecT50-sparse, and Instance-Cholec80-sparse. \robustrevref{2}{1}\revnew{Our process for selecting the frames and sequences for each partition is as follows:}
\begin{enumerate}
    \item Instance-CholecSeg8k: \revmod{We selected} all the 8,080 frames from all the 17 sequences provided in CholecSeg8k, we utilized all the images in CholecSeg8k.
    \item Instance-CholecT50-full: \revmod{We randomly selected}  15 sequences within CholecT50 (CholecT50-full), we utilized all frames provided by CholecT50 for these \revnew{15} sequences \revnew{- 28,317 frames}.
    \item Instance-CholecT50-sparse: For the remaining 35\revnew{(50-15)} sequences in CholecT50(CholecT50-sparse), we sampled these sequences by selecting one frame out of every 30 frames ($\frac{1}{30}$ fps), ensuring a minimum of 50 frames in each sequence \revnew{- 2,681 frames} . 
    \item Instance-Cholec80-sparse: \revmod{After selecting sequences for Instance-CholecSeg8k, Instance-CholecT50 (both full and sparse),} there remained 28 videos from Cholec80, which are not included in CholecT50 and CholecSeg8k, and these videos were sampled at $\frac{1}{30}$ fps, ensuring a minimum of 50 frames in each sequence \revnew{- 2,855 frames} . 
\end{enumerate}
A visual representation of dataset partitions in CholecInstanceSeg can be seen in \figref{fig:our_dataset}.

\robustrevref{1}{6}\revnew{The differences in frame rates for each dataset partition reflect both the design of the parent datasets -- Cholec80 provides videos at 25fps which we then extract at 1fps and CholecT50 provides image sequences at 1fps directly -- and practical considerations for annotating instance segmentation on a large number of frames. Annotating every sequence densely at 1fps is prohibitively expensive. Thus, we adopted a mixed approach, converting existing semantic segmentation to instance segmentation, annotating some sequences \emph{fully} (using all frames available from the parent datasets), and others sparsely (sampling frames at reduced frame rates of $\frac{1}{30}$ fps). This approach allows us to balance annotation density and dataset diversity, ensuring coverage of both densely annotated sequences and a wider range of surgical scenes.}

\robustrevref{2}{1}\revnew{ \figref{fig:recommended_dataset_splits} presents the exact sequences in each partition, represented as VID-XX, where XX is the sequence number. 
}

\subsection*{The Labelling Protocol}
To facilitate the creation of a high-quality instance segmentation dataset for laparoscopic cholecystectomy videos, we established specific rules to ensure accurate and consistent identification and segmentation of surgical instruments. These rules are discussed under three main topics: Tool Categories, Annotation Scope, and Annotation Guidelines.

\subsubsection*{Tool Categories}
In this section, we detail the tool categories selected for annotation and discuss the objects that were deliberately excluded, along with the rationale for these decisions.

We identified seven tool categories for annotation based on our observations from the extracted CholecInstanceSeg sequences:
\begin{enumerate}
    \item Grasper: Standard grasper used for holding tissues.
    \item Bipolar: Bipolar grasper used for grasping and coagulation.
    \item Hook: Monopolar hook used for dissection and coagulation.
    \item Clipper: Clip applier used for applying clips.
    \item Scissors: Monopolar scissors used for cutting tissues.
    \item Irrigator: Suction/irrigation device used for fluid management.
    \item Snare: Laparoscopic snare used for ligation of tissues.
\end{enumerate}

The inclusion of the snare category deviates from the standard six categories (grasper, bipolar, hook, clipper, scissors, irrigator) defined by Cholec80 and CholecT50. We determined that the snare's distinct visual characteristics and unique functionality warranted its classification as a separate tool category. 

\revdel{A visual representation of these tool categories is shown in \figref{fig:tools}}


\subsubsection*{Annotation Scope}
The annotation scope defines the entities included and excluded to maintain a focused and high-quality dataset. While we annotated all laparoscopic tools,  some objects used in conjunction with these tools were excluded to streamline the annotation process:
\begin{itemize}
    \item Excluded Entities: Individual clips from the clipper, surgical needles, surgical sutures, the specimen bag, the camera port, and gauze.
    \item Instrument Ports: We did not annotate the instrument port when no instrument extended from it. However, when an instrument was seen extending from the port, we annotated the instrument port as part of the instrument. This approach aligns with the CholecSeg8k annotation style and addresses the difficulty of distinguishing the exact point where the port ends and the instrument begins. 
    
    \revdel{A clear visual representation of this unique situation with the instrument port can be seen in \figref{fig:instrument_port}}
\end{itemize}


\subsubsection*{Annotation Guidelines}
Our annotation guidelines ensure consistency and accuracy in the labelling process. These guidelines cover various scenarios and provide specific instructions for annotators:
%
%
%
\begin{itemize}
    \item Annotation Quality: We aimed to annotate every visible instrument while maintaining high segmentation quality. Although achieving pixel-perfect annotation for every instance was impractical due to time and budget constraints, we allowed minor imperfections in instrument boundaries, provided that the fundamental characteristics of the instrument and its end-effectors were accurately outlined. Holes within instruments were considered integral parts of the instrument itself, following the standard annotation conventions in popular tool segmentation datasets like EndoVis2017 \cite{endovis2017} and EndoVis2018 \cite{endovis2018}.
    \item Annotation of Hard Cases:  CholecInstanceSeg captures routine surgical procedures, numerous challenging scenarios were encountered. We accounted for twelve specific situations: motion blur, presence of smoke, occlusion by soft transparent tissues or blood, tissues attached to instruments, instruments at the edge of the frame, instruments far from the camera, light reflection, low lighting conditions, bright lighting conditions, lens dirtiness, camera in camera port, and instruments in liquid.
    Specialized instructions and examples were provided to annotators, including checking neighbouring frames for clarification, utilizing the expected shape of the instrument when in doubt, annotating as much of the instrument as possible, and annotating even under low visibility conditions. 
    \revdel{A visual representation of these hard cases can be seen in \figref{fig:hard_cases}.  More detailed information about specialized instructions given to annotators for hard cases is added as supplementary material for this manuscript.} 
\end{itemize}

\subsection*{Annotation Tool and Techniques} \label{sec:annotation_methods}
To provide annotations for CholecInstanceSeg, we needed an annotation tool for performing manual annotations, a method to convert existing semantic segmentation to instance segmentation, and a way to efficiently label a large number of images. This section discusses how we addressed these needs: the annotation tool, converting semantic segmentation to instance segmentation, and semi-automatic annotation.

\subsubsection*{Annotation Tool}\label{sec:annotation_tool}
Data labelling was primarily executed using a minimally customized version of the \href{https://github.com/haochenheheda/segment-anything-annotator}{Segment Anything Annotator tool} \cite{wangsupplementary}, which is open-source software based on the \href{https://github.com/labelmeai/labelme}{LabelMe} annotation project \cite{wada2016labelme}. This tool integrates instrument segmentation functionalities and interactive segmentation capabilities, leveraging the Segment Anything Model (SAM) \cite{kirillov2023segment}. Our customizations included enhancements for navigation, reduction in the size of annotation label files, and the addition of a quality control mode. The customized version is available \href{https://github.com/labdeeman7/my_samannotator}{on github}.

We predominantly utilized point-based SAM prompts for efficient and quick interactive annotation. However, point-prompt-based interactive segmentation often encounters challenges in complex scenarios, such as when instruments and surrounding backgrounds are similar in color, low lighting, glare, blurs, dirty lenses, reflections, and the presence of smoke. In such instances, manual annotation can also be performed using the same tool. Further discussion of challenging scenarios is provided in a later section.

\subsubsection*{Converting Semantic to Instance Segmentation} \label{sec:semantic_to_instance}
One of our data sources, CholecSeg8k, already had semantic segmentation labels. However, we aimed to assign instance IDs to distinguish each instrument instance within an image. To add instance labels and convert existing semantic segmentation labels to instance segmentation, we followed two main steps:

\begin{enumerate}
    \item \textbf{Class-Aware Connected Component Analysis}: We employed connected component analysis for each class separately to identify and segregate different instances within the frames. This technique effectively assigned instance IDs to instruments from different classes and correctly distinguished instances of instruments belonging to the same class that do not overlap in pixel space.
    \item \textbf{Identifying and Addressing Failure Cases}: Class-aware connected component analysis encountered several challenges. A total of 892 frames with these issues were carefully identified and documented. To rectify these problems, we developed and applied several pipelines:
        \begin{enumerate}
            \item \textbf{Occlusions}: When pixels of the same instrument are not connected, class-aware connected component analysis assigns multiple instance IDs to a single instrument. This issue, mainly caused by occlusions at the frame's edge or by tissues, was addressed by combining multiple instances that were actually the same using a Python pipeline.
            \item \textbf{Overlap}: A single instance ID was assigned to different instances of the same instrument class that overlapped in pixel space. We reannotated these frames using our annotation tool.
            \item \textbf{Dataset Noise}: Stray erroneous pixels, incorrect instrument class labels, and missing instrument annotations led to incorrect instance ID assignments. We built pipelines to remove instances caused by dataset noise and correct class labels. Missing instruments were also annotated as required.
        \end{enumerate}
\end{enumerate}

\subsubsection*{Semi-Automatic Annotation}
For the CholecT50-full dataset, which requires annotations for full sequences (nearly 30,000 frames from 15 sequences extracted at 1 fps), we opted for a semi-automatic approach with a human-in-the-loop, utilizing a deep learning model to assist in producing instance segmentation annotations.

\begin{figure}[ht]
  \centering
   \includegraphics[width=0.9\linewidth]{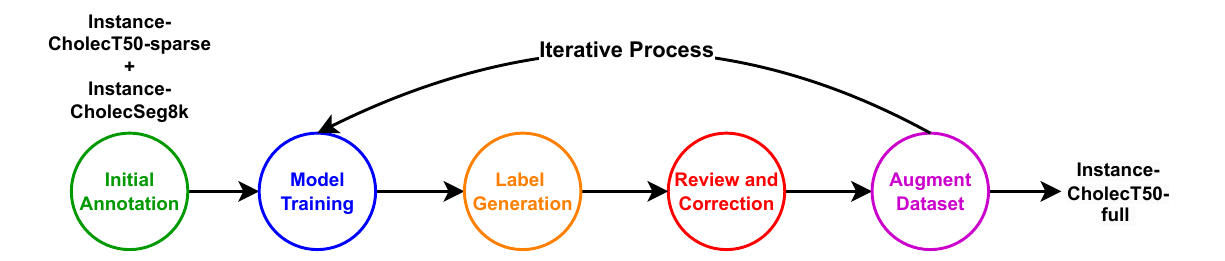}
   \caption{Semi-automatic annotation pipeline. The process begins with initial annotations using Instance-CholecT50-sparse and Instance-CholecSeg8k, followed by model training. The trained model generates labels, which are then reviewed and corrected by human annotators. The corrected annotations augment the training dataset, iterating through the cycle until the final Instance-CholecT50-full dataset is produced.}
   \label{fig:annotation_pipeline}
\end{figure}

\begin{enumerate}
    \item Initial Annotation: We began with initial annotations from the Instance-CholecSeg8k dataset. Since CholecSeg8k only contains two tool categories (Grasper and Hook), which is fewer than the seven tool categories identified for annotation, we performed additional annotations using the annotation tool. We focused on images that included the new instruments. The combined dataset of Instance-CholecSeg8k and these initial annotations provided a foundation to start the semi-automatic pipeline for generating labels.
    \item Model Training:  A deep neural network, the Real-Time Models for Object Detection and Instance Segmentation Large (RTMDet-Ins-l) \cite{lyu2022rtmdet}, was trained using the available annotated datasets (Instance-CholecSeg8k and Instance-CholecT50-sparse). 
    \item Annotation Proposal Generation: Leveraging the trained model, we generated annotations for a larger portion of the unannotated data.
    \item Annotation review and correction: Human annotators reviewed the model-generated annotations and corrected any errors or inaccuracies, ensuring the quality and accuracy of the annotations. \robustrevref{1}{13}\revnew{Minor errors were corrected with the capabilities of the LabelMe labeling tool, while major errors requiring new annotations utilized the point-prompt method of the SAM model.}
    \item Augment training dataset:  The machine learning model was retrained using the expanded training dataset, incorporating the corrected annotations. This iterative process enhanced the model's performance. 
    The model was trained twice: initially, with an annotation set of approximately 11k images and their corresponding labels. The trained models was used to generate labels for about 10k images. Human annotators corrected labels for 5k of these images. The model was then retrained with 21k images, generating labels for another 17k images, with labels for 4k images requiring human annotators for correction.
\end{enumerate}
A visual representation of our semi-automatic approach to annotation can be seen in \figref{fig:annotation_pipeline}.

\subsection*{Annotation Workflow and Procedures}
This section provides an overview of the step-by-step procedure used to create the CholecInstanceSeg dataset, along with detailed insights into the human effort involved in the annotation process. We organize the discussion into two main topics: the step-by-step procedure, and Annotators.

\subsubsection*{Step-by-Step Procedure}

\begin{enumerate}
    \item Annotation of Instance-CholecSeg8k (2 weeks): Converted the CholecSeg8k dataset, which initially contained semantic segmentation labels, into Instance-CholecSeg8k with instance IDs. Quality checks were performed to ensure accuracy. This dataset partition only had two classes and lacked many challenging scenarios present in other datasets.
    \item Creating the Labelling Protocol (3 weeks): Extensive discussions among the annotation team addressed the challenges encountered during the initial annotation attempts of the Instance-CholecT50-sparse partition. These discussions led to the development of a comprehensive labelling protocol to handle hard cases and unique scenarios effectively and consistently.
    \item Annotation of Instance-CholecT50-Sparse (3 weeks): Annotated the Instance-CholecT50 dataset partition using the annotation tool. This occurred in tandem with the creation of the labelling protocol. Quality checks ensured annotations were accurate and consistent with the newly established guidelines.
    \item Annotation of Instance-CholecT50-Full (3 weeks): Using Instance-CholecT50-Sparse and Instance-CholecSeg8k as initial annotations, we employed a semi-automatic pipeline to generate the Instance-CholecT50-Full dataset. This process included iterative model training and corrections to enhance annotation quality.
    \item Annotation of Instance-Cholec80 (1 week): To ensure comprehensive coverage and enhanced diversity for CholecInstanceSeg, we utilized the annotation tool to manually annotate the Instance-Cholec80 partition.
    \item Final Quality Control (2 weeks): Quality control was conducted after the creation of each dataset and again upon finalizing CholecInstanceSeg. This process included several steps to ensure accuracy and consistency. Initially, we performed manual quality control, manually reviewing each image to ensure correct annotations. A second annotator then reviewed a subset of CholecInstanceSeg and performed additional corrections.

    We also leveraged specific properties of CholecInstanceSeg and its relationships with other datasets to enhance quality control. For instance, we used existing tool presence labels from related datasets for cross-validation. Detailed steps for quality control, utilizing these properties and external dataset labels, are extensively discussed in the \hyperlink{sec:quality_control}{quality control section}.
\end{enumerate}

\subsubsection*{Annotators}
The annotation process for creating the CholecInstanceSeg dataset involved a primary annotator responsible for the majority of the annotation tasks and quality control. A secondary annotator, with greater experience and medical qualifications, assisted with annotation and quality control. Both annotators were supported by an expert team to ensure accuracy and consistency in the annotations.

\begin{figure}[ht]
  \centering
   \includegraphics[width=\linewidth]{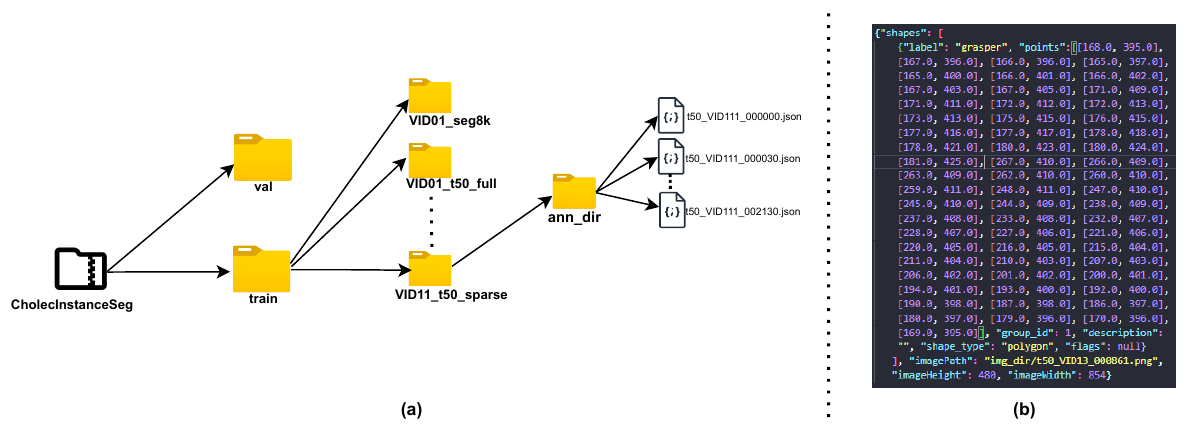}
   \caption{\revmod{(a) Dataset directory structure for CholecInstanceSeg. Each split (e.g. train) contains subdirectories for each sequence, with \revdel{subdirectories for images ($img\_dir$) and} further subdirectories for annotations ($ann\_dir$). (b) Sample JSON annotation file showing the structure of the annotation data, including class labels ("$label$"), polygon information ("$points$")  and instance IDs ("$group\_id$").}}
   \label{fig:file_format}
\end{figure}

\revnew{%
\section*{Data Records}
The CholecInstanceSeg dataset \cite{cholecinstanceseg} is publicly available. Cholec80 and CholecT50 video frames used to generate the dataset can be obtained from the CAMMA research group website(\href{http://camma.u-strasbg.fr/datasets}{http://camma.u-strasbg.fr/datasets}). CholecSeg8k can be obtained at (\href{https://www.kaggle.com/datasets/newslab/cholecseg8k}{https://www.kaggle.com/datasets/newslab/cholecseg8k}).

The dataset is provided as a single archive, $CholecInstanceSeg.zip$, which contains the instance segmentation annotation files in $cholecinstanceseg.zip$ and a metadata file $cholecinstanceseg\_metadata.csv$ that provides metadata information about each annotation file and can be used to link each file to its source dataset. The annotation files are structured into two main directories: train and val, each containing individual video sequence folders. Each sequence folder includes an $ann\_dir$ subdirectory that stores instance segmentation annotations in JSON format. The test directory is withheld for benchmarking purposes and potential competition use, but it is available upon request.
A schematic representation of the dataset folder structure and a sample JSON annotation file are shown in \figref{fig:file_format}

Annotations are formatted following the LabelMe convention. Each shape is listed under the $shapes$ key, with a $label$ field specifying the instrument class, $points$ containing the (x, y) coordinates of the polygon defining the object’s boundary, $group\_id$ indicating the instance ID, $image\_Height$ for the height of image annotated and $image\_Width$ for its width.
}

\section*{Technical Validation}
We provide an in-depth technical validation of the CholecInstanceSeg dataset, demonstrating its reliability and effectiveness for use in research and development of instance segmentation models. We discuss the validation through three main aspects: Data Analysis, training of baseline models, label agreement statistics, and quality control measures.

\subsection*{Data Analysis}
\subsubsection*{Tool Class Distribution}
\figref{fig:instance_tool_spread_chart}(a) shows the distribution of tool classes across the entire dataset. Graspers are the most frequently used tools, followed by hooks. Tools like irrigators, clippers, scissors, and bipolars are used less frequently, with snares being the least used. This indicates a high imbalance in tool class occurrences.

\subsubsection*{Tool Instance Distribution per Frame}
\begin{figure}[ht]
  \centering
   \includegraphics[width=\linewidth]{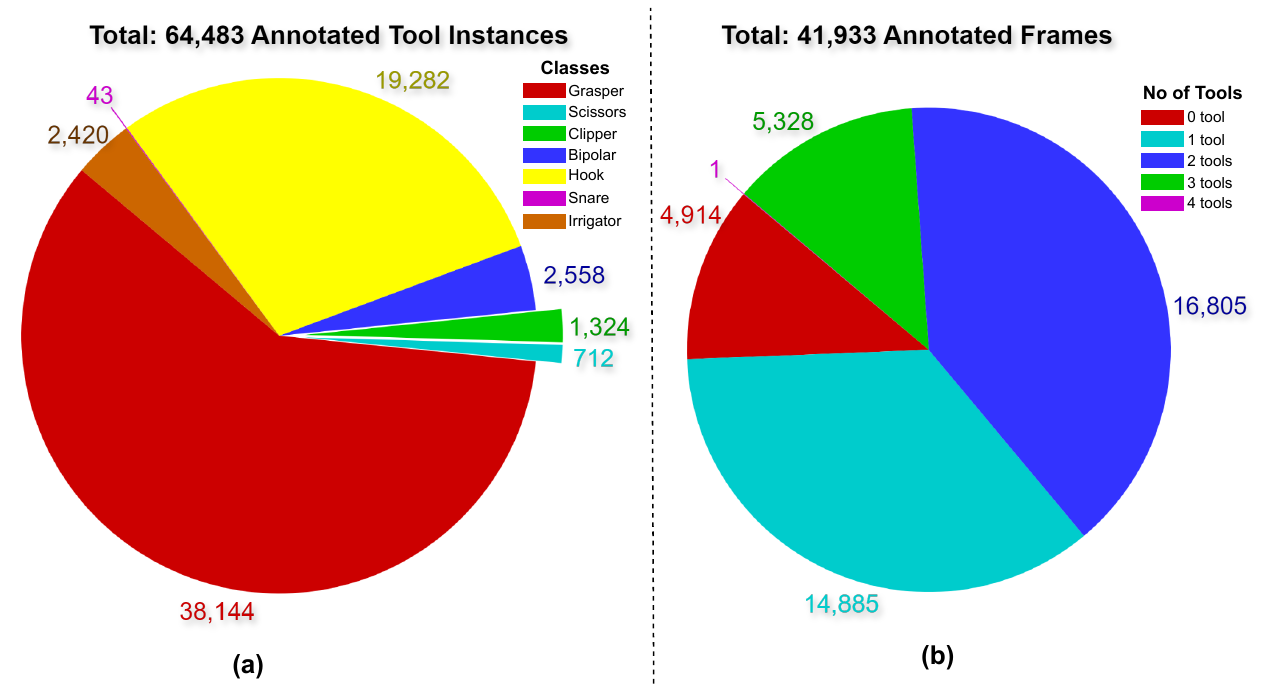}
   \caption{(a) Distribution of each tool class across the dataset. (b) Distribution of the number of tools per frame.}
   \label{fig:instance_tool_spread_chart}
\end{figure}

\figref{fig:instance_tool_spread_chart}(b) illustrates the number of tool instances per frame. The majority of frames contain one or two tools, with a maximum of four tools in a single frame. Approximately 5,328 frames (about one-eighth of the dataset) contain no tools.

\subsubsection*{Partition and Sequence Analysis}
\begin{figure}[ht]
  \centering
   \includegraphics[width=\linewidth]{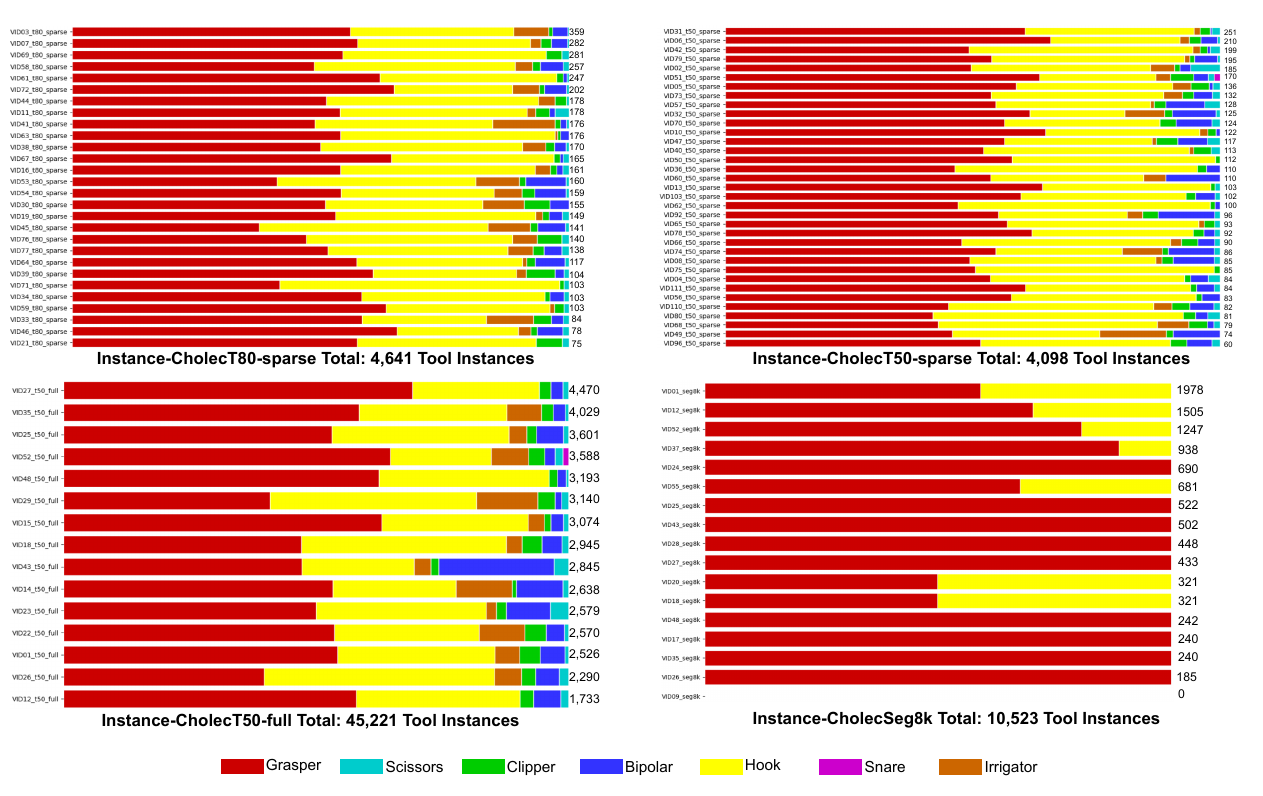}
   \caption{Distribution of tool instances across sequences and partitions in CholecInstanceSeg}
   \label{fig:partition_sequence_chart}
\end{figure}

\tabref{tab:dataset_partition_statistics} provides a summary of the number of sequences, number of frames and amount of annotated tools in each of the dataset partitions.  
\figref{fig:partition_sequence_chart} provides an analysis of tool instances across different sequences and their partitions. Instance-CholecT50-full has the highest number of tool instances, whereas Instance-CholecT50-sparse has the fewest. This figure highlights the imbalance between densely and sparsely annotated sequences and the presence of a sequence with no tools in Instance-CholecSeg8k.

\begin{table}[ht]
  \centering
  \begin{tabular}{|l|l|l|l|l|l|} 
    \hline
      & Inst-CholecSeg8k & Inst-CholecT50-sparse & Inst-CholecT50-full &  Inst-Cholec80-sparse & Total \\
    \hline
    No of seq  & 17(10 shared) & 35 & 15(10 shared) &  28  & 85  \\
    \hline
    No of frames & 8,080 & 2,681 & 28,317  & 2,855  & 41,933  \\
    \hline
    No of tools & 10,523 & 4,098 & 45,221  &  4,641 & 64,483   \\
    \hline    
  \end{tabular}
  \caption{Summary of Dataset Partitions in CholecInstanceSeg.} \label{tab:dataset_partition_statistics}
\end{table}

\subsection*{Training and Evaluation of Baseline Models}

\subsubsection*{Models}
To validate the usability of the CholecInstanceSeg dataset, two baseline instance segmentation models, The Mask R-CNN \cite{he2017mask} and the Mask2Former \cite{cheng2021mask2former} were trained. Since the objective was to establish benchmark performance metrics that future researchers can use for comparison, we decided to utilize more generalizable baseline models instead of optimized tool-specific models like \cite{baby2023forks,zhou2023text}. These models were trained using the training split of CholecInstanceSeg with each model's default hyperparameters. 

\subsubsection*{Recommended Data Splits}
To facilitate effective training, validation, and testing of neural networks for instance segmentation, we divide CholecInstanceSeg into standard training, validation, and testing subsets. This division considers the partitions and shared sequences within the dataset.

For the training dataset, we included the entire Instance-CholecT50-sparse partition, which consists of 35 sequences with 2,681 frames. This partition offers significant diversity in surgical scenes due to its varied sequence content. Additionally, 10 in Instance-CholecSeg8k and the 10 sequences in Instance-CholecT50 which come from the same video (shared) were included in the training dataset to avoid redundancy in the validation and testing subsets. 

The validation dataset contains 3 out of 7 sequences which are unique to Instance-CholecSeg8k. Similarly, from Instance-CholecT50-full, 1 out of 5 sequences which are unique to Instance-CholecT50 were included in the validation dataset. Additionally, half of the sequences from Instance-Cholec80-sparse were added to the validation dataset to bolster its diversity.

The test dataset includes the remaining 4 unshared sequences from Instance-CholecSeg8k and the remaining 4 unshared sequences from Instance-CholecT50-full. The other half of the sequences from Instance-Cholec80-sparse were also included in the test dataset. This division ensures that the test set includes a wide range of scenarios (sequences), providing a comprehensive assessment of model performance.

The final split for the dataset is as follows: the training set contains 55 sequences (26,830 frames), the testing set includes 22 sequences ( \revnew{10,819} \revdel{11,299}  frames), and the validation set comprises 18 sequences (\revnew{4,284} \revdel{3,804} frames ).

\begin{figure}[ht]
\centering
\includegraphics[width=\linewidth]{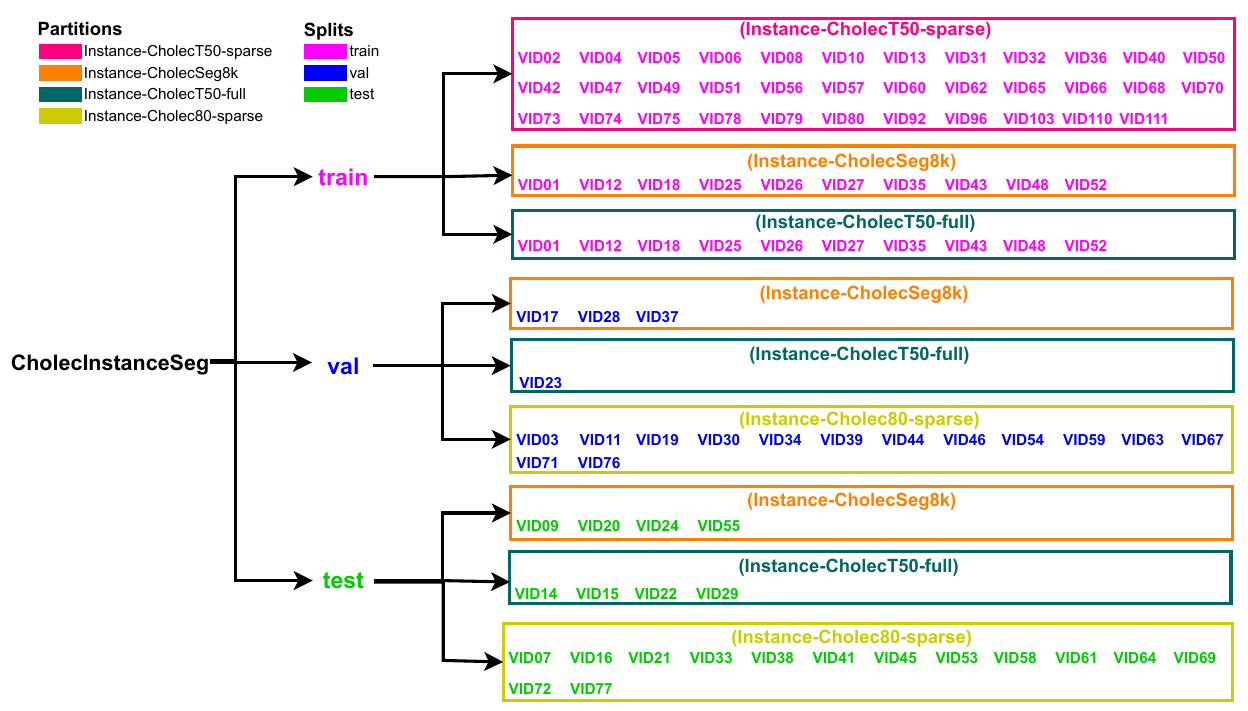}
\caption{Recommended Dataset Splits for CholecInstanceSeg. The diagram shows the distribution of sequences across the training, validation, and testing datasets, highlighting the specific partitions (Instance-CholecT50-sparse, Instance-CholecSeg8k, Instance-CholecT50-full, Instance-Cholec80-sparse) and their corresponding sequences.}
\label{fig:recommended_dataset_splits}
\end{figure}

A visual representation of the recommended dataset split can be seen in \figref{fig:recommended_dataset_splits}. We provide scripts to convert the downloaded zip file into a train, test, and validation format based on these recommended splits.

\subsubsection*{Recommended Metric}
To evaluate our models we used the standard COCO mean average precision metric  (COCO mAP 
 r\revnew{@[0.50:0.95]}).  \revdel{ and a sequence-aware variant, COCO sequence mean average precision (COCO smAP \revnew{@[0.50:0.95]}) }. \robustrevref{2}{4} 

From the dataset analysis, we could tell that our dataset has a class imbalance which is captured by the mean average precision, however, it also has an imbalance among sequences, such that some sequences have a lot of tool instances and images while others do not. 
Hence we wanted to also have a metric that captured how well models are performing on sequences. 

\revdel{COCO smAP generates the average precision per class over each of the sequences separately and then averages this value for the whole dataset. This metric enables the penalization of models that do not generalize to different sequences and scenes.}  \robustrevref{2}{4}

COCO smAP involves a slight modification of the COCO mAP metric. The sequence average precision(sAP) for a class $j$ over the whole dataset ($sAP_j$) in COCO smAP can be written as 
$$
  sAP_j = \frac{\sum{_{s=0}^{N} \int_0^1 p_{seq=s}(j, r) \, dr} }{N}  
$$

where $s$ represents the current sequence, $N$ represents the number of sequences evaluated, and  \( p_{seq=i}(j, r) \) represents the precision at recall \( r \) for class \( j \) over a sequence $s$. \revnew{Essentially $sAP_j$ is calculated by taking the mean of the COCO average precision for that class for each sequence.}

 \robustrevref{2}{4}\revnew{To measure variability in performance across sequences, we introduced the standard deviation of sequence average precision ($\sigma_{sAP_j}$).  
$$
\sigma_{sAP_j} = \sqrt{\frac{1}{N} \sum_{s=0}^{N} \left( \left(\int_0^1 p_{seq=s}(j, r)\right) \, dr - sAP_j \right)^2}
$$

where $s$ represents the current sequence, $N$ represents the number of sequences evaluated, and  \( p_{seq=i}(j, r) \) represents the precision at recall \( r \) for class \( j \) over a sequence $s$. $\sigma_{sAP_j}$ is calculated by taking the standard deviation of the COCO average precision for that class for each sequence.

Finally, the COCO smAP is calculated as a mean of the sequence average precisions across all classes

$$
smAP = \frac{1}{C} \sum_{j=1}^{C} sAP_j
$$
}

We utilize the official COCO implementation present \href{https://github.com/cocodataset/cocoapi/blob/master/PythonAPI/pycocotools/cocoeval.py}{here} for calculating the COCO mAP and the metric COCO smAP metric. \revdel{To determine the COCO smAP metric, we first compute the COCO mAP for each sequence separately and average the results for all sequences.}

\subsubsection*{Results}
The performance results, presented in \revnew{\tabref{tab:baseline_models_map} and \tabref{tab:baseline_models_smap}}, validate the dataset’s effectiveness in training robust instance segmentation models.

 \robustrevref{2}{4}\revnew{Notably, the hook class achieved the highest segmentation performance across both models, likely due to its visually distinct features and higher frequency in the dataset. In contrast, the scissor class exhibited the lowest performance, reflecting its smaller representation and potential challenges in detecting this class. Interestingly, the scissor class demonstrated a higher sequence mAP (smAP) but with significant variability (high standard deviation), suggesting that model performance on this class varies considerably across different sequences.  Increasing the number and diversity of annotated sequences could improve dataset variability and enhance its applicability to a broader range of surgical scenarios, ensuring fewer seen instruments are better recognized by trained models. } 
  \robustrevref{2}{3}



\begin{table}[htp!]
    \centering
    \caption{\robustrevref{2}{4}\revnew{Performance of baseline instance segmentation models on CholecInstanceSeg for the mean average precision metric (mAP) providing the average precision for each class. GR - Grasper, HO - Hook, IR - irrigator, CL - clipper, BI - bipolar, SC - scissors, SN - snare. } }
    \vspace{-1mm}
    \resizebox{\linewidth}{!}{
    \begin{tabular}{l|ccccccc|c}
    \hline
        Method & GR & HO & IR & CL & BI & SC &  SN & mAP  \\ \hline
        Mask-RCNN \cite{he2017mask} & 37.25 & 61.84  & 43.04 & 58.73 & 28.42 & 20.91 & - & 41.70  \\
        Mask2former \cite{cheng2021mask2former} & 61.10  & 82.16 & 68.38 & 81.30 & 48.02 & 27.89 & - & 61.47  \\
     \hline
    \end{tabular}
    }
    \vspace{-4mm}
    \label{tab:baseline_models_map}
\end{table}

\begin{table}[htp!]
    \centering
    \caption{\robustrevref{2}{4}\revnew{Performance of baseline instance segmentation models on CholecInstanceSeg for the sequence mean average precision (smAP) providing the sequence average precision for each class j ($sAP_j$). GR - Grasper, HO - Hook, IR - Irrigator, CL - Clipper, BI - Bipolar, SC - Scissors, SN - Snare. The Standard deviation of the sequence average precision $\sigma_{sAP_j}$ is provided using the ($\pm$) symbol.}}
    \vspace{-1mm}
    \resizebox{\linewidth}{!}{
    \begin{tabular}{l|ccccccc|c}
    \hline
        Method & GR & HO & IR & CL & BI & SC &  SN & smAP  \\ \hline
        Mask-RCNN  & $41.65\pm12.74$ & $64.06\pm9.20$  & $52.50\pm20.09$ & $64.85\pm16.20$ & $37.62\pm20.74$ & $40.38\pm23.54$ & - & $50.18$  \\
        Mask2former & $66.46\pm8.52$  & $88.90\pm7.58$ & $74.51\pm20.35$ & $84.69\pm16.88$ &  $65.20\pm25.36$  & $69.76\pm33.20$  & - & $74.92$  \\
     \hline
    \end{tabular}
    }
    \vspace{-4mm}
    \label{tab:baseline_models_smap}
\end{table}

\subsection*{Label Agreement}
To ensure the accuracy and consistency of the annotations in CholecInstanceSeg, an analysis of label agreement statistics was conducted. Two types of label agreement were calculated. The inter-annotator agreement and manual vs semi-automatic agreement. 

\subsubsection*{Inter-annotator Agreement}
The inter-annotator agreement was assessed using the Panoptic Quality Metric \cite{kirillov2019panoptic}. Two annotators independently annotated a subset of the dataset, and their annotations were compared. The analysis showed a high level of agreement, with a Panoptic Quality score of 91.2, indicating strong consistency between the annotators. The official implementation of the Panoptic Quality metric, available through the \href{https://github.com/cocodataset/panopticapi/tree/master}{panopticapi} repository, was utilized for this evaluation.

\subsubsection*{Manual vs Semi-automatic Agreement}
To validate the consistency of our semi-automatic annotation pipeline against the manual annotation process, we measured the agreement between the two methods using the Panoptic Quality Metric. This comparison was crucial to ensure that the increased speed of the semi-automatic pipeline did not compromise annotation quality. The results demonstrated a high level of agreement, with a Panoptic Quality score of 95.7, indicating that the semi-automatic annotations were consistent with the manual annotations.

\subsection*{Quality Control}\hypertarget{sec:quality_control}{}
Our quality control process involved a series of techniques designed to ensure the accuracy and consistency of the CholecInstanceSeg dataset. In addition to manual quality control and a secondary review by another annotator, we leveraged specific properties of CholecInstanceSeg and tool presence labels from Cholec80. The following steps were employed for quality control:
\begin{enumerate}
    \item checking images with more than 3 tool instances: During annotation, we observed that the most crowded scenes typically included a maximum of three separate tool instances. Therefore, we reviewed all images with more than three tool instances to identify and correct any noise. This process revealed 26 frames with errors due to noise, which were subsequently rectified.
    \item utilizing tool presence labels from Cholec80: Tool presence labels from the Cholec80 parent dataset, although noisy and incomplete, were used for cross-referencing. These labels indicate which tools are present in an image but do not specify the number of tools.
    We cross-referenced our annotations with these tool presence labels to identify frames where an expected tool class from the tool presence labels was missing in CholecInstanceSeg labels. 
    This method identified 267 frames with errors, achieving an approximately 48\% success rate in detecting inaccuracies.   
    \item reducing noise for scenes with tool overlap:  To reduce labelling noise for scenes that include tools occluding other tools, we checked instances that had overlaps. We set a fault tolerance of a 0.1 IoU overlap threshold between tools. All images with intersections exceeding this threshold were reviewed and corrected, resulting in the adjustment of 247 frames. 
\end{enumerate}    

\section*{\revdel{Usage Notes}}
\revdel{CholecInstanceSeg is released under a Creative Commons Attribution-NonCommercial ShareAlike license (CC BY-NC-SA).

Any use or mention of this dataset must include a citation to this paper. If the dataset is utilized in the creation of new works, they should also include citations to this paper. 

This licensing choice aligns with the release license of the Cholec80 dataset, from which our frames are derived, ensuring that the original license is retained and respected.}

\section*{Code Availability}
The relevant code for the use of this dataset is found \href{https://github.com/labdeeman7/cholec_instance_seg}{here}. This repository includes scripts that provide information about class IDs, sequences, and splits. It also contains notebooks for converting between dataset partitions and splits, utilities for converting to the COCO format~\cite{lin2014microsoft}, visualization tools, \revnew{retrieving the input images,} and additional resources. The official implementation of COCO mAP and panoptic quality metric were used and they can be found on the \href{https://github.com/orgs/cocodataset/repositories}{COCO dataset GitHub page}.


\section*{Acknowledgements}
\textbf{Data Sources}: We would like to thank the CAMMA research group at the University of Strasbourg for publishing the data sources without which CholecInstanceSeg would not be possible.
\textbf{Funding Sources} This work was supported by core funding from the Wellcome/EPSRC [WT203148/Z/16/Z; NS/A000049/1] and , Tongji Fundamental Research Funds for the Central Universities. OA is supported by the EPSRC CDT
[EP/S022104/1].
For the purpose of open access, the authors have applied a CC BY public copyright licence to any Author Accepted Manuscript version arising from this submission.

\section*{Author contributions statement}

O.A. wrote the original draft, was the primary annotator, and conducted the experiments and tasks that led to this dataset, K.Z.T. was the secondary annotator for this project, assisted with quality control, and the creation of the annotation guide. Z.Z., C.B., M.S., T.V. were part of the annotation team and mediated ambiguous labels. M.S. and T.V. conceived and supervised this project.  

\section*{Competing interests} 
TV is a co-founder and shareholder of Hypervision Surgical Ltd, London, UK.
The authors declare that they have no other conflict of interest.

\end{document}